\documentclass{article}
\usepackage{xspace}
\usepackage{tabularx}
\usepackage{color,graphicx,soul,graphics}
\usepackage{subcaption}
\usepackage{hyperref}
\usepackage{fullpage}
\newcommand{\csmtool}{CS Materials\xspace}

\AtBeginDocument{%
  }

\begin{document}

\title{Automatic Classification of Pedagogical Materials against CS Curriculum Guidelines}

\author{Erik Saule, Kalpathi Subramanian, Razvan Bunescu\\
  Department of Computer Science\\
  The University of North Carolina at Charlotte\\
  Email:\{esaule, krs, razvan.bunescu\}@charlotte.edu
}
\date{}

\maketitle

\begin{abstract}
Professional societies often publish curriculum guidelines to help
programs align their content to international standards. In
Computer Science, the primary standard is published by ACM and IEEE
and provide detailed guidelines for what should be and could be
included in a Computer Science program.

While very helpful, it remains difficult for program administrators
to assess how much of the guidelines is being covered by a CS
program. This is in particular due to the extensiveness of the
guidelines, containing thousands of individual items. As such, it
is time consuming and cognitively demanding to audit every course to confidently mark
everything that is actually being covered. Our preliminary work
indicated that it takes about a day of work per course.

In this work, we propose using Natural Language Processing techniques
to accelerate the process. We explore two kinds of techniques, the
first relying on traditional tools for parsing, tagging, and embeddings, while the second leverages the power
of Large Language Models. We evaluate the application of these
techniques to classify a corpus of pedagogical materials and show that
we can meaningfully classify documents automatically.


\end{abstract}

\section{Introduction}
The past decade has seen large increases in enrollments in Computer
Science due to a myriad of reasons, including increased opportunities
in the workforce, continued interest in the field and its
applications. This has also created challenges for instructors and
researchers to ensure student success, persistence and degree
completion. Today's students come into CS and related programs with a
wide range of skills and background. At the same time, instructors
need to ensure there is sufficient rigor in the curriculum, so that
the acquired skills will be resilient in the long-term,
despite changes in tools, languages and technologies.

A key point of contention and discussion in Computer Science programs
is deciding what students should know, what we actually teach them,
and whether we should refocus some of our teaching, in line with
current trends. To guide that effort, professional societies have put
out curriculum guidelines to show what could and should be
taught. Most famously, IEEE and ACM have refined Computer Science
curriculum guidelines over the years~\cite{acm-curr_guidelines-2001, acm-curr_guidelines-2013,
  acm-curr_guidelines-2023}.

While these guidelines are quite complete, using them presents a number of challenges. The guidelines are 
extensive, and based on our prior experience, a core issue  is 
the difficulty of carefully mapping even a single course to the curriculum guidelines,
let alone an entire program. We have run workshops to help
audit courses in which the first stage is to classify each module,
each lecture, each assignment against the ACM-IEEE CS curriculum
guidelines from 2013~\cite{acm-curr_guidelines-2013}. The process is tedious and takes 
about a day for an instructor to classify their class. 
However, once that classification is done, it unlocks the ability for the instructor
to get a holistic view of their course, understand how the class is built and 
how it could be improved, compare it to other equivalent courses taught by their colleagues, 
and  can be very helpful in unifying the content of multiple sections of the same course~\cite{goncharow:edupar19, 
goncharow:jpdc21, goncharow:sigcse21}.

In this paper, we investigate the use of automated techniques to
classify the materials of a course against a curriculum guideline. The
purpose is not to entirely automate the classification process,
but rather to accelerate it by enabling instructors to focus
on  the most relevant parts of the guidelines. 
We investigate Natural Language Processing (NLP) tools and design both classical
methods leveraging phrase extraction and word embeddings, and we contrast them
with  using modern pre-trained Large Language Models (LLM).  

Thus \textsl{the research question we address in this work is to assess the effectiveness
of both NLP and LLM based schemes in reducing the initial classification times, which
dominates the effort in aligning course materials to curriculum guidelines}.
We detail a series of methods  and demonstrate that we  can automatically recover 
significant fractions of classifications made by instructors.  This should enable a wider 
adoption of course classification, improving direct education, sharpening
curricular discussion, and simplifying audit processes.


\section{Related Work}

\subsection{Curriculum Guidelines}
ACM and IEEE regularly publish computing curriculum guidelines; the latest 
version is from 2023~\cite{acm-curr_guidelines-2023} and is a refinement from 
2013~\cite{acm-curr_guidelines-2013}. The 2013 guidelines specify a `redefined body
of knowledge, a result of rethinking the essentials necessary for a
Computer Science curriculum'.  The guidelines divide the body of knowledge
into a set of \textit{knowledge areas}; which are further refined into \textit{knowledge units} and 
divided into a set of \textit{topics} and \textit{learning outcomes}.  Learning outcomes 
have three levels, \textit{familiarity, usage} and \textit{assessment}. The 
2023 guidelines~\cite{acm-curr_guidelines-2023} made a number of significant changes; it 
introduced a \textit{knowledge model} to group all the knowledge areas, and as in the 2013
guidelines, knowledge areas consisted of a set of topics and learning outcomes. 
Each knowledge area had a revised Bloom's taxonomy: \textit{Explain, Understand, Apply, 
Evaluate, Develop} and specified \textit{Professional Dispositions}, eg., meticulous, persistence. A competency model was also introduced 
to describe a combination of knowledge, skills and dispositions.

Sub-areas of computing have developed their own standards (parallel 
computing~\cite{tcppcurriculum}, cybersecurity~\cite{CAECD-2019},  
data science~\cite{acmds-curriculum-draft21} and high school CS~\cite{cs-a-14, apcsp-17}) 
which could be used instead or in addition to the ACM/IEEE CS guidelines. Researchers
have also used course syllabi to align with curriculum guidelines~\cite{tungare:sigcse07,becker:sigcse19}, however, it is difficult to derive
reliable insight about the course design and learning objectives in isolation from 
the rest of the course.

\subsection{\csmtool}
We built \csmtool~\cite{goncharow:edupar19, 
goncharow:jpdc21, goncharow:sigcse21}, an interactive and online 
system  for CS instructors and CS education researchers.
\csmtool enables mapping CS educational materials (lectures, assignments,
videos, etc.) to concepts defined by curriculum guidelines (topics and learning outcomes). 
The system currently supports the ACM/IEEE 2013 guidelines~\cite{acm-curr_guidelines-2013} and
the NSF/IEEE-TCPP Parallel and Distributed Computing 2012 guidelines~\cite{tcppcurriculum}. 
It provides friendly user interfaces for instructors to classify their course content and 
and the ability to structure a class 
in a hierarchy of class-modules-materials to help with organization and structural analysis.
It supports analysis features to help an instructor understand the 
structure of their class~\cite{mcquaigue:eduhpc23} or a CS education researcher 
to understand more general trends~\cite{goncharow:sigcse21}. The system also 
has search features to enable one to find materials that correspond to particular 
concepts, either through an explicit search or as recommendations for a particular class 
or module. The system has been used by dozens of instructor and feature about 1700 materials.

\subsection{NLP Methods in CS Education}
NLP and ML methods have a long history of automating various 
tasks in education, as illustrated by the yearly Workshop on the Innovative Use of NLP for 
Building Educational Applications (BEA) that started at NAACL in 2003~\cite{ws-2003-hlt-naacl}.
Pre-LLM methods relied on regression or classification models trained on 
vectors of manually engineered features and lexical similarity measures, as in the 
short answer grading system from~\cite{mohler:acl11}. The emergence of LLMs, together 
with the breadth of knowledge encoded in their parameters and their increasingly 
sophisticated linguistic, reasoning, and theory of mind skills, have  led 
to numerous applications in education~\cite{bea-2023-innovative,bea-2024-innovative}. 
As student support, LLMs have been used to generate hints for math word 
problems~\cite{sridhar:aied23} and programming exercises~\cite{leinonen_using_2023}, 
to generate code explanations~\cite{macneil:sigcse23}, or to Socratically guide 
students~\cite{al-hossami_can_2024}. On the instructor’s side, LLMs have helped
with authoring learning materials~\cite{macneil:sigcse23} and learning objectives~\cite{sridhar_harnessing_2023}, or generating multiple choice questions 
and short-answer grading~\cite{doughty:ace24}.

\section{Course Classification Process}
\label{sec:class-process}

\paragraph{Initial Classification.}
The entire ACM/IEEE 2013 guidelines are part of the \csmtool system. The system provides
an interactive interface for a user to classify each material (lecture slides, quizzes, 
videos, etc.) by selecting the most relevant entries in the guidelines. A list interface allows
users to navigate to the knowledge areas of interest and select those items that
match the material's topics and learning outcomes. Typically, the initial materials take
time to classify, as the user needs to get familiar with the guidelines, but becomes 
more efficient as more materials are classified. From our experience, an entire 
course can take upto a day to classify. It is also the 
most time consuming part of the classification process.  On the positive side, users get 
a good understanding of the curriculum guidelines and their value towards good 
course design.

\paragraph{Reconciliation view.}
Once the initial classification is completed, \csmtool provides a
matrix view of the classification with the guideline entries (topics,
outcomes) as rows and the learning materials as
columns. This view helps the instructor to review the entire
classification of their course and harmonize the classification accross
materials.  This process usually takes less than 20 minutes. However,
this view only contains those guideline entries that were
selected for at least one material of the course.

\paragraph{Coverage and Alignment Analysis.}
In addition, \csmtool provides a radial view of the \textsl{entire
  ACM/IEEE guideline hierarchy}.  This view contains not only highlighted items
selected by the user for their course, but also shows items in
the same knowledge unit. The user has  an opportunity to review 
entries in the classification that are not marked, and possibly
uncover topics and outcomes that are covered in their course
but may have been misclassified. This process usually takes less than 15 minutes.

In addition, the radial view provides a visualization of the course coverage and has features
to compare two equivalent courses  with the ability
to highlight differences, that could potentially lead to course changes or discussion between
the two course instructors.

\paragraph{Seeding with course templates.}
An important advantage of the \csmtool is the ability to start with an existing classification
in the system and use that as a template for a new course.
This works when there is a similar and well classified
course in the system since the user can skip the initial classification step and 
use the remaining steps to refine the classification of their course.

\paragraph{Takeaways.}
The initial classification step is the most time consuming part of the
course classification process, especially for a user who is not familiar with 
the curriculum guidelines.
The remaining steps are for reviewing the classification and are of minimal
effort. Thus, our focus is on making the initial classification
process more efficient  through the use of automated techniques.

\section{Classification using Classical NLP}

The core problem we are tackling is: given a course material
as a PDF file and the curriculum guidelines in a machine readable file format
(e.g., JSON tree), produce a small (say, 20) ranked list of topics and learning outcomes that are  covered by the input material. Henceforth, we shall use the term {\it category} to refer to the topics and learning outcomes specified in the curriculum guidelines.

\subsection{Preprocessing}

The materials that we classify are PDF files. We use a PDF processing library to extract the
text. We evaluate pypdf~\cite{pypdf-doc} and
pymupdf~\cite{pymupdf-doc}. These libraries can leave ligatures in the
text that we replaced automatically, for instance double ``ﬀ'' create
ligatures.

Then, our classical NLP methods use base noun phrases (bNP) from each
category and the text of the material. We define a base noun phrase as a maximal sequence of nouns (N) that may be preceded by one ore more adjectives (J), i.e.  "(J.*)*(N.*)+" sequences. The text is tokenized and tagged using NLTK~\cite{bird2006nltk} and its Part-of-Speech~\cite{marquez-2000-postag} 
tagging model. 

\subsection{Exact Phrase Matching}

Our first set of methods match base noun phrases exactly. For each category
and for each bNP  in the document, we will count how
many times that bNP appears in the text of the material.
We  return the occurrence frequencies of the bNPs in decreasing order
in the document.  We call this method \texttt{count-unweighted}.
Note that because we bNPs are maximal sequences, 
``dynamic programming'' is not a match to ``efficient dynamic
programming''.

A variation of this, we call \texttt{count-weighted}, is to be more
sensitive to the occurrence frequency of the phrases; in other words,
normalize the impact of a bNP by the number of classification
entries that bNP appears in. The rationale for this approach is
that a rarely occurring phrase might be more relevant than a more
commonly occurring phrase.

\subsection{High Dimension Word Embedding}
A problem of the previous method is that it relies on exact phrase
matching. There may be equivalent phrases that are not recognized. For
instance ``breadth-first-search'' and ``BFS'' refer to the same algorithm, and
a ``graph'' is sometimes referred to as a ``network''.

Word embeddings were essentially designed to solve this problem. They
map words or phrases to a semantic high dimensional vector space.
We use a pretrained embedding table called `glove-wiki-gigaword-300'
from Gensim~\cite{rehurek_lrec}, a general purpose embedding of words into 300
dimensions.  For each phrase, we compute its representation by taking the average
representation of all the words. Words that do not
appear in the embedding table are ignored. 


We compute distances between two embeddings using their cosine
similarity. We will take distances as $1-abs(cosine)$. We consider two
phrases matched if their distance is below a threshold of $0.3$. We
consider two variants for matching a phrase from a document to a
phrase in a classification item: either we consider \texttt{all}
phrase matching (within the distance threshold) or we consider only
the \texttt{best} match (as long as it is within the threshold).

When a phrase from a document matches a phrase from a category, we use the same \texttt{unweighted} and \texttt{weighted}
variants from the exact phrase matching methods. This gives us four variants:
\texttt{embedding-weighted-all}, \texttt{embedding-weighted-best},
\texttt{embedding-unweighted-all}, \texttt{embedding-unweighted-best}.

\section{Classification using Large Language Models}

\subsection{Basic LLM Methods}

While word embeddings are better than lexical phrases at capturing semantic similarity, the meaning captured in a word embedding is still context independent, which is problematic when the a word's actual meaning depends on the context. This means that the word "dynamic" when used in "dynamic programming" (a type of algorithm) is associated the same embedding as when used in "dynamic partitioning" (a memory management technique). In contrast, embeddings produced by Large Language Models (LLM) integrate contextual information through the self-attention mechanism~\cite{vaswani_attention_2017}. As such, their autoregressive generations captures deep contextual clues in text, which is important for our classification task.

The simplest idea is to give the LLM the whole text document
(extracted using PyMuPdf) and the whole ontology of the curriculum
guidelines.  The query is expected to return a list of the IDs of the
curriculum guidelines that were matched by the document. The LLama-3.3 model used in our system supports a maximum of 128K input tokens, which is often insufficient for such a large request. It
is likely that commercial LLMs like Gemini will be  able to process such a large
query, but we did not evaluate this option.

The queries in the previous method are too long because the curriculum guidelines
are quite extensive. For reference, the JSON encoding of the entire hierarchy
(knowledge areas, knowledge units, topics and learning outcomes) is about 630KB.
So we used a simpler approach by asking for 
topics and outcomes one at a time. For a document, we made one
query per category, asking whether it covers
a specific category, and to provide a ``yes''/``no'' answer. This resulted in a 
binary classifier for each category, generating a complete list of
classified categories. We call this method \texttt{llm-binary}.

In preliminary tests, we could tell that the LLM was often returning
too many categories; further investigation revealed that the binary decision
prevented the LLM from expressing nuances. So we rephrased the question 
to ask the LLM to provide a score from 0 (not a match at all) to 5 (definitely a match). 
This allows sorting the categories by matching quality. We call
this method \texttt{llm-5point}.

\subsection{Runtime Optimizations}
The downside of using an LLM as a one-category-at-a-time classifier is
that it results in about 2700 queries (number of categories in the ACM
2013 guidelines) per document.  This is computationally expensive, as
it takes time to query the system, which can
dampen the utility of the tool. Secondly, on a local server, the
queries can degrade performance for other users. On a cloud system,
it can incur large financial charges (per minute or per
token).

Thus, we modified our application to use asynchronous queries to the
LLM.  While this significantly improve the time to classify a document
thanks to concurrency, it does not decrease the computational
cost. Since this does not impact the classification generated, we will
only report runtimes in the experimental section.

Secondly, we grouped queries together. Each query to the LLM is provided five categories 
from the guidelines and asked to rate all five of them. We call this
method \texttt{llm-5pointbatch}.

\subsection{Leveraging Additional Context}

Surprisingly, the \texttt{llm-5pointbatch} seemed to provide
better classification than the non-batched experiment. We hypothesize
that this happens because the categories are no
longer considered in isolation, but rather in the context of
the other categories in the same query. We designed two more methods 
to provide more context to the LLM to improve accuracy.
First, we designed \texttt{llm-5point-context}, which provides not only
the category but also the knowledge
area and the knowledge unit that the category is part of.

Second, we designed \texttt{llm-prune-5point-context}, that takes it a
step further. First, during initialization, the LLM 
generates a summary of each knowledge unit in a few sentences,
based on the topics and outcomes in that unit. Then for each
document, the summary is presented to the LLM and is asked whether
any categories in the knowledge unit could be a match. Only if the LLM answers
in the positive do we ask to rate each category in that knowledge
unit, one at a time, using a 5 point scale. This should
both decrease the number of queries by pruning parts of the guidelines,
and improve the quality of the response.

\section{Experiments}

\begin{figure}[t]
  \center
  \includegraphics[width=.48\linewidth]{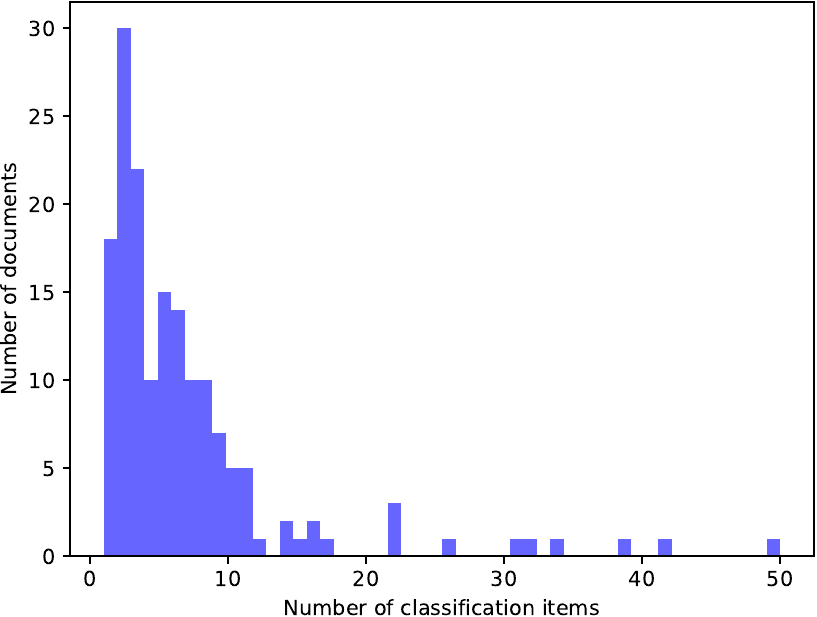}
  \caption{Number of categories in documents}
  \label{fig:hist_classification}
\end{figure}

\subsection{Dataset}

For our experiments, we are using the public documents in
\csmtool that have been classified against the
ACM/IEEE CS 2013 Curriculum guidelines. We only keep the categories for which
a PDF file is attached, for a total of 170 documents. The documents in
the dataset are from six different CS1 courses, three different Data Structures and Algorithms
courses, one Architecture course, one Networking course, and one Operating Systems course.
Because the documents were classified by
different instructors, there is variability in the
classifications. Figure~\ref{fig:hist_classification} shows that most
documents have been classified with fewer than 10 categories,
while some have more than 30. The average is $6.65$.

In comparison, the curriculum guidelines has 2790 categories (including
the headings for Knowledge Areas, and Knowledge Units). This tells us
that the classification of a document is very sparse. We configured
the methods to return their top-20 classifications.

\subsection{Classical NLP Methods}

\begin{table}
  \center
  \begin{tabular}{|l|r|r|}
\hline
 & \multicolumn{2}{c|}{Recall} \\
Method &  pydpdf & pymupdf \\
\hline\hline
count-unweighted & 11.35\% & 10.67\% \\\hline
count-weighted & 13.27\% & 14.09\% \\\hline
embedding-unweighted-allmatch & 20.89\% & 20.35\%  \\\hline
embedding-unweighted-bestmatch & 13.92\% & 14.34\%  \\\hline
embedding-weighted-allmatch & 18.24\% & 16.58\%  \\\hline
embedding-weighted-bestmatch & 15.95\% & 17.21\%  \\\hline
\end{tabular}

  \caption{Recall of classic NLP method}
  \label{tab:classicNLP-pypdf}
  \label{tab:classicNLP-pymupdf}
\end{table}

We start by looking into the performance of the classic NLP
technique to get a sense of baseline
behaviors. Table~\ref{tab:classicNLP-pypdf} presents the results of
the different classic NLP methods on our dataset. We consider the
results using the pypdf text extractor first.

The embeddings based method performed better than the counting based
method. The counting based method recovered about 11-13\% of the
classification that the instructors performed. The embedding based
methods on the other hand performed better, recalling in the range of
13-21\% of the instructor classification.

\begin{figure}
  \begin{subfigure}[t]{0.47\linewidth}
    \includegraphics[width=\linewidth,page=6]{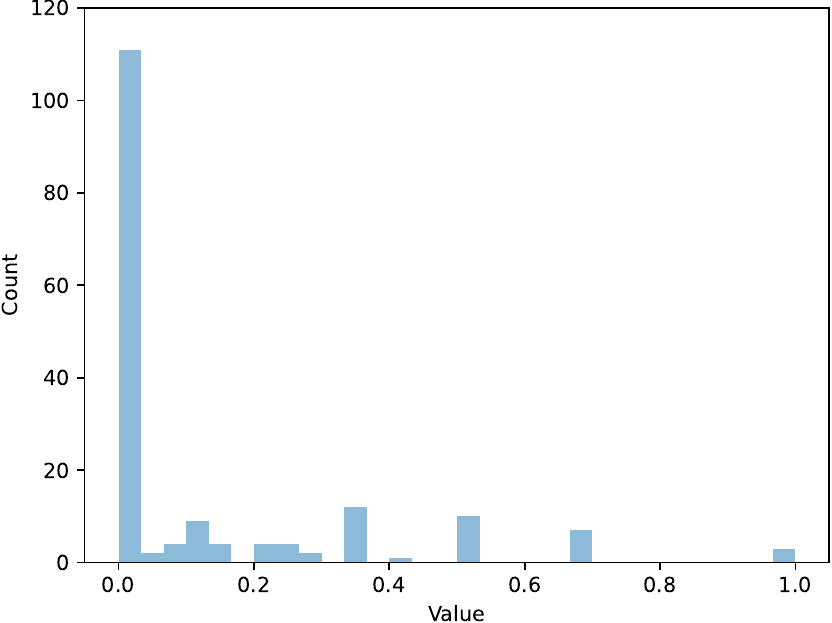}
    \caption{\texttt{count-weighted}}
  \end{subfigure}
  \begin{subfigure}[t]{0.47\linewidth}
    \includegraphics[width=\linewidth,page=7]{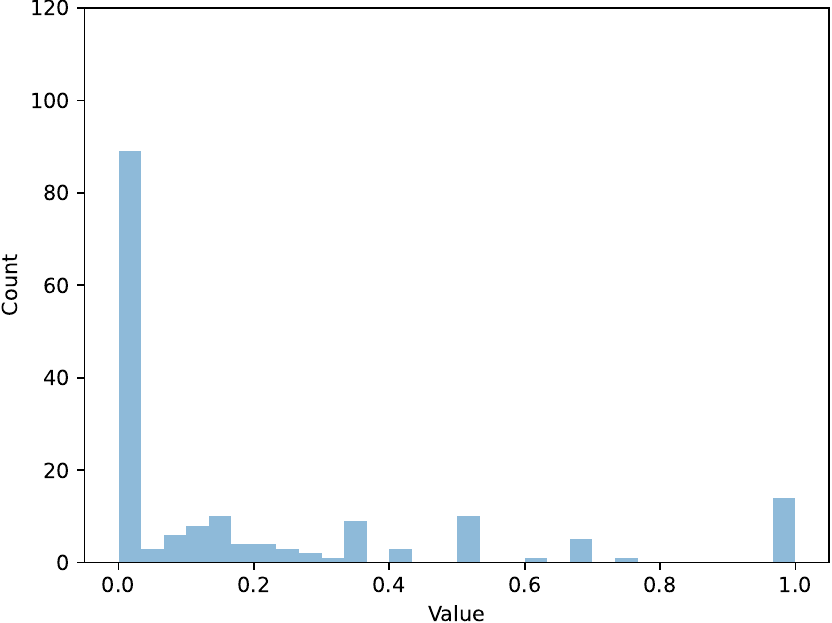}
    \caption{\texttt{embedding-unweighted-all}}
  \end{subfigure}
  \caption{Distribution of recalls}
  \label{fig:recall-dist}
\end{figure}

The best counting method is \texttt{count-weighted} and the embedding method is
\texttt{embedding-unweighted-all}. Figure~\ref{fig:recall-dist} shows
the distribution of the recall per document for both methods. The
embedding based methods has a better distribution: it has more
documents with a recall of 1, fewer documents with a recall of 0. The
intermediate values of recall are roughly distributed equally.

\begin{figure}
  \center
  \includegraphics[width=.45\linewidth,page=5]{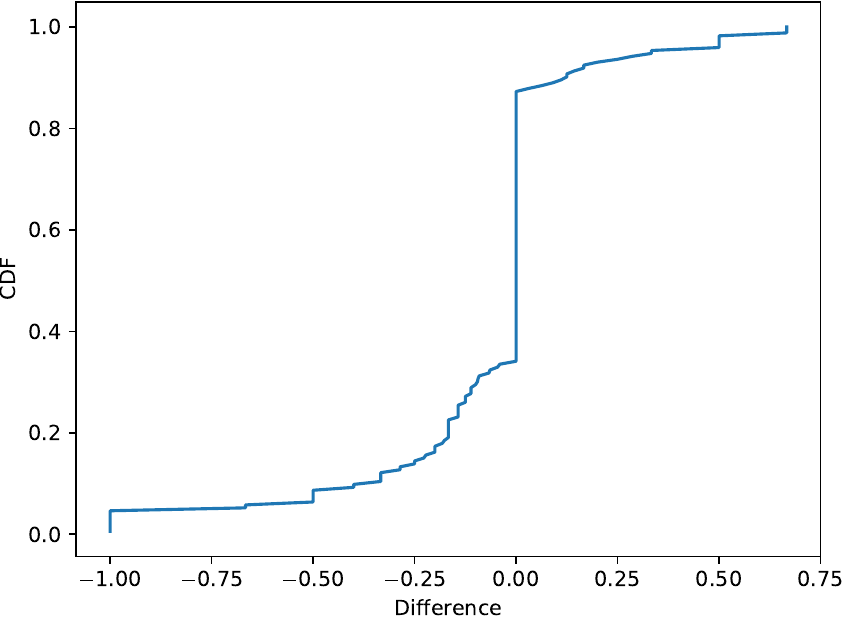}
  \caption{Cumulative density function of the recall difference between \texttt{count-weighted} and \texttt{embedding-unweighted-all}. Positive values indicate \texttt{count-weighted} does better.}
  \label{fig:recall-diff}
\end{figure}

The results indicate that the embedding methods do better  in the aggregate. But do they perform better on all documents?  Figure~\ref{fig:recall-diff} answers that
question by presenting the distribution over all documents of the
difference of the recall achieved by \texttt{count-weighted} and
\texttt{embedding-unweighted-all}.

For about half the documents, the difference is 0, indicating both
algorithms do as well. But \texttt{count-weighted} does better for
15\% of the documents, and \texttt{embedding-unweighted-all} does
better for 35\% of the documents. This indicates that there is
room for improvement in these methods: relevant information
is present in the documents, but the methods are not connecting the
dots.

A manual inspection of the documents showed a couple of issues. The text
extraction from PDF is not always the best. In documents with figures, some
of the text is mangled, while those with code,  variable
names get mangled in the text. Jargon is often not recognized correctly by the 
Part-of-Speech algorithm~\cite{marquez-2000-postag}  and words at
beginning of phrases  are often recognized as
proper nouns.

Table~\ref{tab:classicNLP-pymupdf} also presents recall values for a
different PDF text extractor: pymupdf. The difference in performance
between the two PDF processors is small, usually within 1.5\% in
recall. However, we do see manually that the text seems to be a bit
better extracted by PyMuPdf which also supports more file formats.

\subsection{LLM Results}

The results of the different versions of the LLM based methods are
presented in Table~\ref{tab:llm-pymupdf}. \texttt{llm-binary} performs
worse than the best embedding based method, recalling only 18\% of the
instructor classification. Manually inspecting the output of the
method reveals that the method matches
hundreds of classification items. As we only retain 20 of the items  and the
decisions are binary, the technique is unable  to distinguish between great matches 
and good matches.

Our \texttt{llm-5point} method is able to overcome that problem and
recover 28\% of the instructors classifications. Manual inspection of the
output of the method shows that many classification items are ranked
with a score of 5, and there are usually more than 20 of them. 
This method ranks with a score of  5 far fewer items that the \texttt{llm-binary}
classified as ``yes'' (about three times fewer).

\begin{table}
  \center
  \begin{tabular}{|l|r|}
\hline
Method & Recall \\
\hline\hline
llm-binary & 18.59\%  \\\hline
llm-5point & 28.62\%  \\\hline
llm-5pointbatch & 42.30\%  \\\hline
llm-5pointcontext & 42.53\%  \\\hline
llm-prune-5pointcontext & 47.65\%  \\\hline
\end{tabular}

  \caption{Recall of LLM-based methods}
  \label{tab:llm-pymupdf}
\end{table}

Manually inspecting the output of \texttt{llm-5point} shows that the
method sometimes lack context. For instance, a topic in ``Discrete
Structure'' is ``Functions'', as in \textit{mathematical functions}. But a lecture about \textit{function calls} is deemed as matching the
topic.

The \texttt{llm-5point-batch} and \texttt{llm-5point-context} methods
add contextual information and perform about the same, achieving a
recall of 42\%. The neighboring topics for \texttt{llm-5point-batch},
and the knowledge area and knowledge units for
\texttt{llm-5point-context} seem to help disambiguate the topics and
outcomes. These methods rank with a score of 5 about 30\% fewer
entries than \texttt{llm-5point}.

\begin{figure}
  \center
    \includegraphics[width=.45\linewidth,page=6]{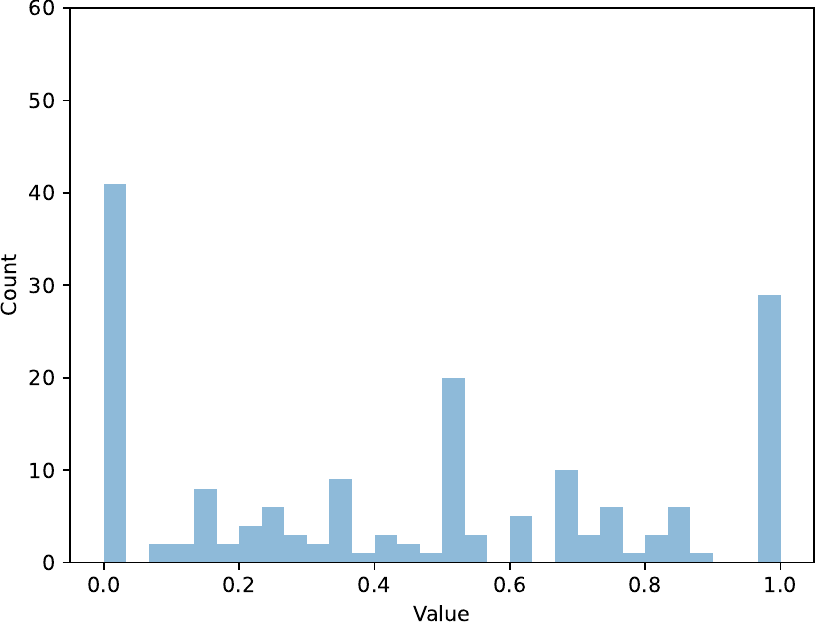}
    \caption{Recall distribution of \texttt{llm-prune-5pointcontext}}
    \label{fig:llm-prune-recall-dist}
\end{figure}

Finally, the \texttt{llm-prune-5point-context} method performs the best. The pruning of entire knowledge units helps
preventing many misclassifications. It is able to recall 47\% of the
instructor classifications. We provide in
Figure~\ref{fig:llm-prune-recall-dist} the distribution of recall per
document for \texttt{llm-prune-5point-context}. We can see that the
performance has dramatically improved from the classical NLP
methods. About 30 of the documents are perfectly recovered by the
method, yet 33 documents do not see their classification
recovered at all.

\subsection{Runtimes}

We next discuss the runtime of the different methods to grasp
their computational costs. Note that the comparisons are
fundamentally unfair. The classical NLP techniques are run single-threaded on a laptop
and the code is written in Python without 
performance optimization. The LLM based methods are running
on a server equipped with four NVIDIA A5000. Yet we believe that the
difference in orders of magnitude highlights the difference in 
computational costs.

The classic NLP techniques based on phrase counts took about 20
minutes to classify all documents. The method based on embeddings took
about an hour.  The methods based on LLM were more expensive. 
Without leveraging concurrent requests, the binary and 5 point rating methods 
took 37 hours. Batching 5 classifications in a single query reduced the
runtime to 30 hours.

The main gain in time came from using concurrent requests. We used a
particular document (id:1599) as a benchmark, and classifying with one
query at a time took 570 seconds, with 10 concurrent queries
took 200 seconds, and using 50 concurrent queries dropped
it further to 127 seconds. After that the performance tapers as
the server saturates, and using 90 concurrent queries took 120
seconds. We settled on using 40 queries at a
time which provides most of the performance benefit without saturating
the server. Using asynchronous processing for the entire dataset dropped the 
processing time to 5 hours 40 minutes. But this does not reduce the volume of 
computation, it just schedules the calculation more intelligently.

\texttt{llm-prune-5point-context} actually does reduce the volume of
calculations. It prunes the number of queries per document from 2700 to
about 550, which represents a 5-fold decrease. The experimental time decreased to 4 hours, 
however it should be noted that the implementation does not perfectly leverage concurrent queries.

\section{Discussion}

\subsection{Analysis of error}

The \texttt{llm-prune-5point-context} method has a recall at 20 of
47\%. While it is not a bad result, in order to judge how useful that
classification might be, we need to understand what happens when it makes incorrect classifications. We manually combed
through the 33 documents for which the method had no match to the
instructors' classification. We looked at  the document, 
the instructor's classification, and  the output and logs of
the method.

In a few cases, the documents contain almost no text; it
is essentially a set of pictures with basic captions. So no
relevant text get extracted, and  automatic classification
was not possible. A second set of 
documents seemed to be misclassified by the instructor. The
documents do not seem to match the instructor's classification. We
believe that the wrong files were uploaded during the input process.

The vast majority of documents (about 20 of them) seem to have been
misclassified for a similar reason: The document was classified by the
instructor based on how the document fits in their class, rather than
based on what the document directly talks about. For example, a
lecture on linked list used in a Data Structures course was classified
by the instructor using list and data structure categories.
But the document itself describes the implementation of
linked list in Java and how to structure classes and function
calls. So the method identified Object-Oriented Programming related
categories first, and ranked the data structure categories
second. But these were past the top-20 and thus not
returned. The documents were classified correctly in the top-40
classification items but not in the top-20. And the top-20 is filled
is low-level categories while the instructor's classification  
focused on high-level categories.

\subsection{On using it in a course classification exercise}

If the NLP techniques could automatically and always come up with
precisely the right topics and learning outcomes, that would be
method of choice. But in practice, they are not quite that accurate. The methods
need to return quite a few items to get the right coverage. So
in practice there will be a need for human review.

We believe that the methods that we presented would integrate well in
the course classification process that we have used in prior workshops
and described in Section~\ref{sec:class-process}.  We can replace the
initial classification phase by pre-seeding it with the output of the
automatic classification method. By only showing the entries of the
curriculum guidelines recommended by the method and the entries that
surround the guidelines, we can enable the instructors to
focus on the entries that are most relevant.

Note that the methods do not need to identify all the entries correctly
for each material. The reconciliation and the coverage views will
still be used in harmonizing the classification across materials in
the same course. And the coverage view can still identify missing
entries. So overall, the initial classification needs to be close to
correct but does not need to be perfect.

The second question is whether using the automatic classifier would
significantly reduce the time to classify and entire
class. Classifying each document automatically takes a couple minutes each,
which could be batched in a background process. The work
of the instructor would be to review the classification. But with the
method focusing attention on a couple dozen classification entries
only spanning a few knowledge units, the classification of one document
would only take a few minutes. And a full class could be
initially classified within a couple of hours. The matrix review and
coverage review would then take about the same amount of time as before,
and together, within an hour.

Therefore, we believe that one could classify an entire class within
half a day using automated classification, and save about half a day
compared to the manual classification process.

\section{Limitations}

There are some  limitations to our proposed methods.
The dataset we are using is from real instructors doing classification for the 
purpose of understanding their own courses. It is a small dataset that  does 
not necessarily follow the best practices followed by Machine Learning and Natural Language 
Processing communities. In particular,
a well defined task requires the design of initial annotation guidelines that are updated with feedback 
from a reconciliation step where human annotators discuss their disagreements. Once annotation guidelines
are finalized, two or more annotators would label the same set of examples, in order to compute
a measure of inter-annotator agreement (ITA).

Also, the methods we present are an initial evaluation of these
tools. There may be other embedding methods that could yield better
results. We could fine-tune embeddings to specialize them to Computer
Science for better results. Similarly
we are using an off-the-shelf LLM model in our evaluation and not more
powerful commercial models running on much larger systems. One could
also probably achieve better results by fine-tuning the language
model to do well on our classification task, or through the use of in-context learning examples. 
We are performing some amount of
prompt engineering but more can certainly be done.

We do not see these issues as being critical because we are not trying
to fully automate the course classification process. Rather we have
shown the potential of our methods for improving the instructor
classification process with experimental results. However, we have not
actually conducted controlled experiments involving instructors that
prove that the tools would make the process faster.

\section{Conclusion}

In this paper, we have done some initial exploration on using NLP tools to help 
classify course materials against the ACM 2013 curriculum guidelines. We showed that 
Large Language Models can be used to recover and match 47\% of entries selected by the 
instructor, using the top-20 suggestions from the method. Error analysis shows that this is a very 
conservative estimate of performance, e.g., due to human annotation mistakes or lack of signal in the input 
materials, as such actual classification performance is likely to be higher.
We expect that these techniques can
significantly accelerate the task of classifying our courses, and ease the path of the
the CS education community towards better analysis of course content.

While these initial results are encouraging, we need to confirm them by
designing human studies where instructors would classify
their courses in a system and collect their feedback on the
process. We expect that simplifying the classification of courses
against curriculum standards would improve the reusability of content
and enable smarter analysis of what we teach to our students.

\section*{Acknowledgment}
This material is based upon work supported by the National Science Foundation under Grant No. DUE-2142381.

\bibliographystyle{abbrv}
\bibliography{paper}

@inproceedings{vaswani_attention_2017,
	title = {Attention is {All} you {Need}},
	volume = {30},
	url = {https://proceedings.neurips.cc/paper_files/paper/2017/file/3f5ee243547dee91fbd053c1c4a845aa-Paper.pdf},
	booktitle = {Advances in {Neural} {Information} {Processing} {Systems}},
	publisher = {Curran Associates, Inc.},
	author = {Vaswani, Ashish and Shazeer, Noam and Parmar, Niki and Uszkoreit, Jakob and Jones, Llion and Gomez, Aidan N and Kaiser, \Lukasz and Polosukhin, Illia},
	editor = {Guyon, I. and Luxburg, U. Von and Bengio, S. and Wallach, H. and Fergus, R. and Vishwanathan, S. and Garnett, R.},
	year = {2017},
}

@inproceedings{rehurek_lrec,
      title = {{Software Framework for Topic Modelling with Large Corpora}},
      author = {Radim {\v R}eh{\r u}{\v r}ek and Petr Sojka},
      booktitle = {{Proceedings of the LREC 2010 Workshop on New
           Challenges for NLP Frameworks}},
      pages = {45--50},
      year = 2010,
      month = May,
      day = 22,
      publisher = {ELRA},
      address = {Valletta, Malta},
      note={\url{http://is.muni.cz/publication/884893/en}},
      language={English}
}

@inproceedings{bird2006nltk,
  title={NLTK: the natural language toolkit},
  author={Bird, Steven},
  booktitle={Proceedings of the COLING/ACL 2006 interactive presentation sessions},
  pages={69--72},
  year={2006}
}

@Article{marquez-2000-postag,
  author = 	 {Llu\'{i}s M\`{a}rquez and  Llu\'{i}s Padr\'{o} and Horacio Rodr\'{i}guez },
  title = 	 {A Machine Learning Approach to POS Tagging},
  journal = 	 {Machine Learning},
  year = 	 {2000},
  OPTkey = 	 {},
  OPTvolume = 	 {},
  number = 	 {39},
  pages = 	 {59-91},
  month = 	 apr,
  OPTnote = 	 {},
  OPTannote = 	 {}
}

@Manual{pymupdf-doc,
  title = 	 {PyMuPDF documentation},
  OPTkey = 	 {},
  author = 	 {Artifex Software},
  OPTorganization = {},
  OPTaddress = 	 {},
  OPTedition = 	 {},
  OPTmonth = 	 {},
  year = 	 {2025},
  URL = {https://pymupdf.readthedocs.io/en/latest/},
  OPTnote = 	 {},
  OPTannote = 	 {}
}

@Manual{pypdf-doc,
  title = 	 {pypdf documentation},
  OPTkey = 	 {},
  author = 	 {Mathieu Fenniak},
  OPTorganization = {},
  OPTaddress = 	 {},
  OPTedition = 	 {},
  OPTmonth = 	 {},
  year = 	 {2025},
  URL = {https://pypdf.readthedocs.io/en/stable/},
  OPTnote = 	 {},
  OPTannote = 	 {}
}

@proceedings{ws-2003-hlt-naacl,
    title = "Proceedings of the {HLT}-{NAACL} 03 Workshop on Building Educational Applications Using Natural Language Processing",
    year = "2003",
    OPTurl = "https://aclanthology.org/W03-0200/"
}

@inproceedings{mohler:acl11,
author = {Mohler, Michael and Bunescu, Razvan and Mihalcea, Rada},
title = {Learning to grade short answer questions using semantic similarity measures and dependency graph alignments},
year = {2011},
isbn = {9781932432879},
publisher = {Association for Computational Linguistics},
address = {USA},
booktitle = {Proceedings of the 49th Annual Meeting of the Association for Computational Linguistics: Human Language Technologies - Volume 1},
pages = {752–762},
numpages = {11},
location = {Portland, Oregon},
series = {HLT '11}
}

@inproceedings{sridhar_harnessing_2023,
	title = {Harnessing {LLMs} in {Curricular} {Design}: {Using} {GPT}-4 to {Support} {Authoring} of {Learning} {Objectives}},
	shorttitle = {Harnessing {LLMs} in {Curricular} {Design}},
	url = {http://arxiv.org/abs/2306.17459},
	doi = {10.48550/arXiv.2306.17459},
	urldate = {2024-04-14},
	publisher = {arXiv},
	author = {Sridhar, Pragnya and Doyle, Aidan and Agarwal, Arav and Bogart, Christopher and Savelka, Jaromir and Sakr, Majd},
	month = jul,
	year = {2023},
	address = {Tokyo, Japan},
	booktitle = {Proceedings of the Workshop on Empowering Education with LLMs—the Next-Gen Interface and Content Generation}
}

@inproceedings{doughty:ace24,
author = {Doughty, Jacob and Wan, Zipiao and Bompelli, Anishka and Qayum, Jubahed and Wang, Taozhi and Zhang, Juran and Zheng, Yujia and Doyle, Aidan and Sridhar, Pragnya and Agarwal, Arav and Bogart, Christopher and Keylor, Eric and Kultur, Can and Savelka, Jaromir and Sakr, Majd},
title = {A Comparative Study of AI-Generated (GPT-4) and Human-crafted MCQs in Programming Education},
year = {2024},
isbn = {9798400716195},
publisher = {Association for Computing Machinery},
address = {New York, NY, USA},
url = {https://doi.org/10.1145/3636243.3636256},
doi = {10.1145/3636243.3636256},
booktitle = {Proceedings of the 26th Australasian Computing Education Conference},
pages = {114–123},
numpages = {10},
location = {Sydney, NSW, Australia},
series = {ACE '24}
}

@inproceedings{al-hossami_can_2024,
	address = {New York, NY, USA},
	series = {{SIGCSE} 2024},
	title = {Can {Language} {Models} {Employ} the {Socratic} {Method}? {Experiments} with {Code} {Debugging}},
	isbn = {979-8-4007-0423-9},
	shorttitle = {Can {Language} {Models} {Employ} the {Socratic} {Method}?},
	url = {https://doi.org/10.1145/3626252.3630799},
	doi = {10.1145/3626252.3630799},
	urldate = {2024-04-14},
	booktitle = {Proceedings of the 55th {ACM} {Technical} {Symposium} on {Computer} {Science} {Education} {V}. 1},
	publisher = {Association for Computing Machinery},
	author = {Al-Hossami, Erfan and Bunescu, Razvan and Smith, Justin and Teehan, Ryan},
	month = mar,
	year = {2024},
	pages = {53--59},
}

@inproceedings{leinonen_using_2023,
	address = {New York, NY, USA},
	series = {{SIGCSE} 2023},
	title = {Using {Large} {Language} {Models} to {Enhance} {Programming} {Error} {Messages}},
	isbn = {978-1-4503-9431-4},
	url = {https://dl.acm.org/doi/10.1145/3545945.3569770},
	doi = {10.1145/3545945.3569770},
	urldate = {2023-10-29},
	booktitle = {Proceedings of the 54th {ACM} {Technical} {Symposium} on {Computer} {Science} {Education} {V}. 1},
	publisher = {Association for Computing Machinery},
	author = {Leinonen, Juho and Hellas, Arto and Sarsa, Sami and Reeves, Brent and Denny, Paul and Prather, James and Becker, Brett A.},
	month = mar,
	year = {2023},
	pages = {563--569},
}

@inproceedings{macneil:sigcse23,
author = {MacNeil, Stephen and Tran, Andrew and Leinonen, Juho and Denny, Paul and Kim, Joanne and Hellas, Arto and Bernstein, Seth and Sarsa, Sami},
title = {Automatically Generating CS Learning Materials with Large Language Models},
year = {2023},
isbn = {9781450394338},
publisher = {Association for Computing Machinery},
address = {New York, NY, USA},
url = {https://doi.org/10.1145/3545947.3569630},
doi = {10.1145/3545947.3569630},
booktitle = {Proceedings of the 54th ACM Technical Symposium on Computer Science Education V. 2},
pages = {1176},
numpages = {1},
location = {Toronto ON, Canada},
series = {SIGCSE 2023}
}

@inproceedings{sridhar:aied23,
  author = {Sridhar, Pragnya and Doyle, Aidan and Agarwal, Arav and Bogart, Christopher and Savelka, Jaromír and Sakr, Majd},
  editor = {Moore, Steven and Stamper, John C. and Tong, Richard Jiarui and Cao, Chen and Liu, Zitao and Hu, Xiangen and Lu, Yu and Liang, Joleen and Khosravi, Hassan and Denny, Paul and Singh, Anjali and Brooks, Christopher},
  ee = {https://ceur-ws.org/Vol-3487/paper9.pdf},
  pages = {139-150},
  publisher = {CEUR-WS.org},
  series = {CEUR Workshop Proceedings},
  title = {Harnessing LLMs in Curricular Design: Using GPT-4 to Support Authoring of Learning Objectives.},
  url = {http://dblp.uni-trier.de/db/conf/aied/llm2023.html#SridharDABSS23},
  volume = 3487,
  year = 2023
}

@proceedings{bea-2024-innovative,
    title = "Proceedings of the 19th Workshop on Innovative Use of NLP for Building Educational Applications (BEA 2024)",
    editor = {Kochmar, Ekaterina  and
      Bexte, Marie  and
      Burstein, Jill  and
      Horbach, Andrea  and
      Laarmann-Quante, Ronja  and
      Tack, Ana{\"i}s  and
      Yaneva, Victoria  and
      Yuan, Zheng},
    month = jun,
    year = "2024",
    address = "Mexico City, Mexico",
    publisher = "Association for Computational Linguistics",
    url = "https://aclanthology.org/2024.bea-1.0/"
}

@proceedings{bea-2023-innovative,
    title = "Proceedings of the 18th Workshop on Innovative Use of NLP for Building Educational Applications (BEA 2023)",
    editor = {Kochmar, Ekaterina  and
      Burstein, Jill  and
      Horbach, Andrea  and
      Laarmann-Quante, Ronja  and
      Madnani, Nitin  and
      Tack, Ana{\"i}s  and
      Yaneva, Victoria  and
      Yuan, Zheng  and
      Zesch, Torsten},
    month = jul,
    year = "2023",
    address = "Toronto, Canada",
    publisher = "Association for Computational Linguistics",
    url = "https://aclanthology.org/2023.bea-1.0/"
}

@inproceedings{mcquaigue:eduhpc23,
  author =   {Matthew Mcquaigue and Erik Saule and Kalpathi
                  Subramanian and Jamie Payton},
  title =    {Data-Driven Discovery of Anchor Points for PDC
                  Content},
  OPTcrossref =  {},
  OPTkey =   {},
  booktitle =    {Proceedings of SC23 Workshops (SC-W); EduHPC},
  doi =      {10.1145/3624062.3624099},
  year =     {2023},
  type =     {workshop},
  url =      baseurl:esaule:publis#{eduhpc23-MSSP.pdf},
  slides =   baseurl:esaule:slides#{saule23-eduhpc.pdf},
  OPTnote =  {},
  OPTannote =    {}
}

@inproceedings{goncharow:sigcse21,
  author = {Goncharow, Alec and Mcquaigue, Matthew and Saule, Erik and
                  Subramanian, Kalpathi and Payton, Jamie and Goolkasian,
                  Paula},
  title = {Mapping Materials to Curriculum Standards for Design,
                  Alignment, Audit, and Search},
  year = {2021},
  isbn = {9781450380621},
  publisher = {Association for Computing Machinery},
  address = {New York, NY, USA},
  doi = {10.1145/3408877.3432388},
  booktitle = {Proceedings of the 52nd ACM Technical Symposium on
                  Computer Science Education},
  pages = {295–301},
  numpages = {7},
  location = {Virtual Event, USA},
  series = {SIGCSE '21},
  url = {http://webpages.uncc.edu/~esaule/public-website/papers/sigcse21-GMSSPG.pdf},
  youtube = {https://youtu.be/w9cbVlf2CUw},
  type = {conference}
}

@book{apcsp-17,
    Author = {{College Board}},
    Publisher = {College Board},
    Title = {AP Computer Science Principles, Including the Curriculum 
		Framework},
    Year = {Fall 2017}}

@book{cs-a-14,
    author = {{College Board}},
    publisher = {College Board AP},
    title = {Computer  Science A: Course Description},
    year = {Fall 2014},
	OPTaddress = {New York, NY, USA},
	note = {\url{https://apcentral.collegeboard.org/pdf/ap-computer-science-a-course-description.pdf}}
}

@book{acm-curr_guidelines-2013,
    author = {{The Joint Taskforce on Computing Curricula: Association for Computing Machinery (ACM), IEEE Computer Society}},
    publisher = {ACM/IEEE Computer Society},
    title = {Computer Science Curricula 2013: Curriculum Guidelines for 
		Undergraduate Degree Programs in Computer Science},
    year = {2013},
	OPTaddress = {New York, NY, USA},
	note = {\url{https://www.acm.org/binaries/content/assets/education/cs2013_web_final.pdf}}
}

@book{acm-curr_guidelines-2023,
author = {Kumar, Amruth N. and Raj, Rajendra K. and Aly, Sherif G. and Anderson, Monica D. and Becker, Brett A. and Blumenthal, Richard L. and Eaton, Eric and Epstein, Susan L. and Goldweber, Michael and Jalote, Pankaj and Lea, Douglas and Oudshoorn, Michael and Pias, Marcelo and Reiser, Susan and Servin, Christian and Simha, Rahul and Winters, Titus and Xiang, Qiao},
title = {Computer Science Curricula 2023},
year = {2024},
isbn = {9798400710339},
publisher = {Association for Computing Machinery},
note = {\url{https://dl.acm.org/doi/pdf/10.1145/3664191}},
address = {New York, NY, USA}
}

@book{acm-curr_guidelines-2001,
    author = {{The Joint Taskforce on Computing Curricula: IEEE Computer Society, Association for Computing Machinery}},
    publisher = {ACM/IEEE Computer Society},
    title = {Computing Curricula 2001 Computer Science},
    year = {2001},
	note = {\url{http://www.acm.org/binaries/content/assets/education/curricula-recommendations/cc2001.pdf}}
	}

@inproceedings{tungare:sigcse07,
 author = {Tungare, Manas and Yu, Xiaoyan and Cameron, William and Teng, GuoFang and P{\'e}rez-Qui\~{n}ones, Manuel A. and Cassel, Lillian and Fan, Weiguo and Fox, Edward A.},
 title = {Towards a Syllabus Repository for Computer Science Courses},
 OPTbooktitle = {Proceedings of the 38th SIGCSE Technical Symposium on Computer Science Education},
 booktitle = {Proc. of ACM SIGCSE},
 series = {SIGCSE '07},
 year = {2007},
 location = {Covington, Kentucky, USA},
 pages = {55--59},
 numpages = {5},
 OPTpublisher = {ACM},
 OPTaddress = {New York, NY, USA},
}

@TechReport{tcppcurriculum,
  author =	 {{NSF/IEEE-TCPP Curriculum Working Group}},
  title =	 {{NSF/IEEE-TCPP} Curriculum Initiative on Parallel
                  and Distributed Computing : Core Topics for
                  Undergraduates},
  institution =	 {CDER},
  year =	 {2012},
  note =	 {Available at
      \url{http://www.cs.gsu.edu/~tcpp/curriculum/sites/default/files/NSF-TCPP-curriculum-version1.pdf}}
}

@TechReport{acmds-curriculum-draft21,
  author =	 {{ACM Data Science Task Force}},
  title =	 {{Computing Competencies for Undergraduate Data
                  Science Curricula}},
  year =	 {2021},
  month =	 jan,
  institution =	 {ACM},
  OPTurl =
                  {https://www.acm.org/binaries/content/assets/education/curricula-recommendations/dstf_ccdsc2021.pdf}
}

@TechReport{CAECD-2019,
  author =	 {{National Security Agency}},
  title =	 {{Centers of Academic Excellence in Cyber Defense
                  (CAE-CD) -- 2019 Knowledge Units}},
  institution =	 {NSA},
  year =	 {2018},
  note =	 {\url{https://www.iad.gov/NIETP/documents/Requirements/CAE-CD_2019_Knowledge_Units.pdf}}
}

@inproceedings{becker:sigcse19,
 author = {Becker, Brett A. and Fitzpatrick, Thomas},
 title = {{What Do CS1 Syllabi Reveal About Our Expectations of Introductory Programming Students?}},
 booktitle = {Proceedings of the 50th ACM Technical Symposium on Computer Science Education},
 series = {SIGCSE '19},
 year = {2019},
 isbn = {978-1-4503-5890-3},
 location = {Minneapolis, MN, USA},
 pages = {1011--1017},
 numpages = {7},
 note = {\url{http://doi.acm.org/10.1145/3287324.3287485}},
 doi = {10.1145/3287324.3287485},
 acmid = {3287485},
 publisher = {ACM},
 address = {New York, NY, USA},
}

@article{goncharow:jpdc21,
  author = {Goncharow, Alec and Mcquaigue, Matthew and Saule, Erik
                  and Subramanian, Kalpathi and Goolkasian, Paula and Payton,
                  Jamie},
  title = {CS-Materials: A System for Classifying and Analyzing
                  Pedagogical Materials to Improve Adoption of Parallel and
                  Distributed Computing Topics in Early CS Courses},
  journal = {Journal of Parallel and Distributed Computing},
  doi = {10.1016/j.jpdc.2021.05.014},
  year = {2021},
  url = {http://webpages.uncc.edu/~esaule/public-website/papers/jpdc21-GMSSGP.pdf},
  optannote = {}
}

@article{goncharow:edupar19, 
	title={Classifying Pedagogical Material to Improve Adoption of Parallel and Distributed Computing Topics}, 
	url={http://par.nsf.gov/biblio/10091591}, 
	journal={9th NSF/TCPP Workshop on Parallel and Distributed Computing Education (EduPar-19)}, 
	author={Goncharow, Alec and boekelheide, Anna and Mcquaigue, Matthew and Burlinson, David and Saule, Erik and Subramanian, Kalpathi}, 
	year={2019}, 
	month={May}
}

@STRING{baseurl:esaule:publis={http://moais.imag.fr/membres/erik.saule/papers/} }

@STRING{baseurl:esaule:publis={http://bmi.osu.edu/~esaule/public-website/paper/} }

@STRING{baseurl:esaule:publis={http://webpages.uncc.edu/~esaule/public-website/papers/} }

@STRING{baseurl:esaule:slides={http://webpages.uncc.edu/~esaule/public-website/slides/} }
\end{document}